\documentclass[runningheads]{llncs}
\usepackage{amsopn}
\usepackage{graphicx}
\usepackage{hyperref}
\usepackage{verbatim} 
\usepackage[normalem]{ulem}
\graphicspath{ {./img/} }
\usepackage{xcolor}
\usepackage{multirow}
\usepackage{subcaption}
\begin{document}

\title{Panoptic Segmentation in Industrial Environments using Synthetic and Real Data}

\titlerunning{Panoptic Segmentation using Synthetic and Real Data}

\author{Camillo Quattrocchi\inst{1} \and
Daniele Di Mauro\inst{1} \and \\
Antonino Furnari\inst{1,2} \and
Giovanni Maria Farinella\inst{1, 2}}

\authorrunning{C. Quattrocchi et al.}

\institute{FPV@IPLAB, DMI - University of Catania, Italy \and Next Vision s.r.l. - Spinoff of the University of Catania, Italy}

\maketitle 

\begin{abstract}
      
      Being able to understand the relations between the user and the surrounding environment is instrumental to assist users in a worksite. For instance, understanding which objects a user is interacting with from images and video collected through a wearable device can be useful to inform the worker on the usage of specific objects in order to improve productivity and prevent accidents.
      Despite modern vision systems can rely on advanced algorithms for object detection, semantic and panoptic segmentation, these methods still require large quantities of domain-specific labeled data, which can be difficult to obtain in industrial scenarios. Motivated by this observation, we propose a pipeline which allows to generate synthetic images from 3D models of real environments and real objects. The generated images are automatically labeled and hence effortless to obtain. Exploiting the proposed pipeline, we generate a dataset comprising synthetic images automatically labeled for panoptic segmentation. This set is complemented by a small number of manually labeled real images for fine-tuning. Experiments show that the use of synthetic images allows to drastically reduce the number of real images needed to obtain reasonable panoptic segmentation performance.
      
\end{abstract}

\section{Introduction}
    In recent years, the increasing development of wearable devices has sparked new interest on First Person Vision~\cite{betancourt2015evolution}. Among the different application contexts, First-Person Vision is a must-go approach in industrial environments, where it can be useful to improve productivity and increase safety. Possible applications include operator training, anticipation of the imminent use of a potentially dangerous object, and predictive maintenance via the estimation of machinery usage.
    Developing systems able to tackle the aforementioned tasks requires large amounts of data to develop, train, and evaluate computational models. Despite datasets such as Ego4D\footnote{https://ego4d-data.org/}, EPIC-KITCHENS~\cite{9084270}, EGTEA-Gaze+~\cite{li2021eye} and MECCANO~\cite{Ragusa_2021_WACV} have been recently proposed, they only offer a partial answer to the need of data in industrial contexts. Indeed, these datasets only partially consider industrial scenarios~\cite{Ragusa_2021_WACV} and even if they do, the environment-specific nature of industrial applications often requires models to be trained on domain-specific data acquired on purpose. Moreover, these datasets do not contain labels for the assessment of semantic segmentation models.
    
    This lack of data limits the development of algorithms in industrial settings because building a dataset in such contexts is a challenging task. Indeed, designing a proper data acquisition and labeling protocol requires time and expertise. Furthermore, the data acquisition process can be long and requires the availability of a diverse set of subjects, environments and representative situations. Additionally, once the images of the dataset have been acquired, the required labeling stage (usually a manual procedure) can become a long and costly process. Data acquisition is even more challenging in the industrial domain where it is easy to run into privacy and confidentiality issues which may slow down or prevent the collection of appropriate data. These negative aspects related to the generation of real image datasets can be outrun using synthetic data~\cite{9412149}: once a 3D model of the target environment and objects has been acquired, the data generation and labeling can be automated. This allows to greatly increase the speed and reduce the costs of the data collection process. While the generated data can be directly used to train the model, one of the main concerns is to reach a degree of photorealism of the synthetic data to achieve good results when the trained model is tested on real world data. The amount of photorealism depends mainly on the quality of the 3D model, the software used for the generation and the computational capabilities of the adopted 3D acquisition hardware. If properly designed, synthetic data also allows to increase diversity, for instance including rare events which would be difficult to observe during the acquisition of a training set composed of real images.
    
    The drop in performance of computer vision algorithms between synthetic and real data is known as \emph{domain shift}~\cite{Csurka2017}. When available, real data can be used to fill the gap between the real and synthetic domains through fine-tuning or more complex domain adaptation techniques~\cite{pasqualino2021unsupervised}. Even in the presence of real data, synthetic data allows to greatly reduce the amount of real data needed to reach good results, which brings benefits in terms of the reduction of costs in the acquisition and labeling stages.
    
    In this paper, we investigate the impact of synthetic data for the development of domain-specific applications in industrial environments. As a primary task, which can be useful in many downstream applications, we propose to study panoptic segmentation~\cite{Kirillov_2019_CVPR}, which consists in identifying the main semantic elements in the scene, including both structural parts, such as walls, and object instances such as tools and equipment. Specifically, we study the suitability of training a panoptic segmentation approach using a large amount of labeled synthetic data and very small amount of labeled real data. Since both the label taxonomy and the data are inherently domain-specific in industrial settings, we propose a pipeline to generate synthetic data compliant to a specific real industrial environment. We hence generate and publicly release a dataset~\footnote{The dataset is available at \url{https://iplab.dmi.unict.it/ENIGMA_SEG/}.} containing synthetic images generated from the 3D model of a real industrial space, as well as labeled real images collected in the same space.
    
    In sum, the contributions of this work are twofold: (1) a pipeline for the generation of a synthetic dataset specifically designed for the panoptic segmentation task in an industrial site; (2) a labeled dataset generated using the proposed pipeline. The dataset consists of labeled real and synthetic data where each frame has four different annotations: Semantic Label, Panoptic Image, Instance Annotation, Panoptic Annotation.
    
    To explore the usefulness of the proposed pipeline and synthetic data, we carry out a benchmark of Panoptic Segmentation~\cite{Kirillov_2019_CVPR}. The benchmark analyzes the performances of the baseline model proposed in~\cite{Kirillov_2019_CVPR} when the synthetic data is used for model training and variable amount of real data are used to fine-tune the model. The results show that the synthetic data can be useful to obtain a good Panoptic Segmentation model.
    
    The remainder of this paper is structured as follows: in Section~\ref{related} we discuss related works; Section~\ref{dataset} presents the proposed dataset generation pipeline; Section~\ref{benchmark} reports and discusses the results; Section~\ref{conclusion} concludes the paper.
    
\section{Related Work}
\label{related}    
    The work in this paper is related to the computer vision tasks of Semantic Segmentation, Instance Segmentation, Panoptic Segmentation and to the use of synthetic datasets for learning purposes.

    \subsection{Virtual Dataset Generation}
    
    The use of synthetic data to train machine learning models has become increasingly popular. Ragusa et al.~\cite{9412149} collected both a synthetic and a real dataset related to a cultural site to study the performance of semantic segmentation models in the presence of varying amounts of synthetic and real data. Chang et al.~\cite{Matterport3D} proposed to obtain synthetic images of an indoor environment from a Matterport scan to train a semantic segmentation model, highlighting the difficulty to collect large amounts of real data in order to train scene understanding algorithms. Orlando et al.~\cite{orlando2019image} introduced a tool for the generation of synthetic tours in a virtual environment. The developed tool allows the navigation of an agent within a simulated environment. 
    
    Di Benedetto et al.~\cite{di2019learning} acquired a synthetic dataset using a Game Engine: the authors managed to obtain good results for object detection even with a small amount of real data to be used for fine-tuning. The works in~\cite{richter2017playing,fabbri2021motsynth,krahenbuhl2018free,hu2019sail,hu2021sail} demonstrate the power of generating synthetic data automatically annotated for multiple tasks, such as optical flow estimation, semantic instance segmentation, object detection and tracking, object-level 3D scene layout estimation and visual odometry.
    
    Similarly to these works, we propose to use synthetic data to tackle a supervised learning task. Differently from the previous approaches, we propose a pipeline that allows the generation of synthetic data aligned to real industrial spaces.

    \subsection{Semantic, Instance and Panoptic Segmentation}
    
    Semantic segmentation is the task of assigning a class to each pixel in an image~\cite{lateef2019survey}. Semantic segmentation aims to give more information than object detection: in addition to the position of objects, shape and geometry can also be derived. To solve this task, various approaches have been proposed in the last years such as encoder-decoder, dilated convolutional models, multiscale and pyramid networks~\cite{7803544,lan2020global,Zhao_2017_CVPR}.
    
    One of the first works using deep learning to tackle semantic segmentation is known as SegNet~\cite{7803544}, a deep network which belongs to the category of encoder-decoder models. DeepLab, presented in~\cite{chen2017deeplab}, became a seminal work with the introduction of ``atrous convolution'', a powerful tool in dense prediction tasks, and ``atrous spatial pyramid pooling (ASPP)'' to robustly segment objects at multiple scales. Chang et al.~\cite{Zhao_2017_CVPR} proposed a semantic segmentation network that exploits global context information using a pyramid pooling layer.
    
    The instance segmentation task consists in detecting and segmentating objects in images. The tasks includes the localization of specific objects and the inference of whether pixels belong to each object in order to discern an instance from another. Chen et al. in~\cite{chen2019hybrid} proposed a cascade method for instance segmentation, taking full advantage of the relationship between detection and segmentation. Tian et al.~\cite{tian2021boxinst} proposes a very powerful method capable of performing instance segmentation using only the bounding boxes at training time.
    
    Kirillov et al.~\cite{Kirillov_2019_CVPR} introduced the panoptic segmentation task. The authors of~\cite{Kirillov_2019_CVPR} introduced the difference between ``thing'' and ``stuff'' classes and how semantic segmentation and instance segmentation, typically distinct tasks, can be unified. The paper also proposed a new metric, Panoptic Quality (PQ), to measure performance.
    Hwang et al.~\cite{hwang2021exemplar} proposed a variant of panoptic segmentation, called open-set panoptic segmentation which aims to recognize also the classe unknown at training time.
    
    In this work, we consider Panoptic Segmentation as a reference task to test the value of the generated synthetic dataset. We choose this task because it enables a general understanding of the scene at a semantic level, including structural elements (``stuff'') and object instances (``things''), which can be useful in industrial contexts.

\section{Dataset}
\label{dataset}
    
    \begin{figure}[t!]
            \centering
            \includegraphics[width=0.9\textwidth]{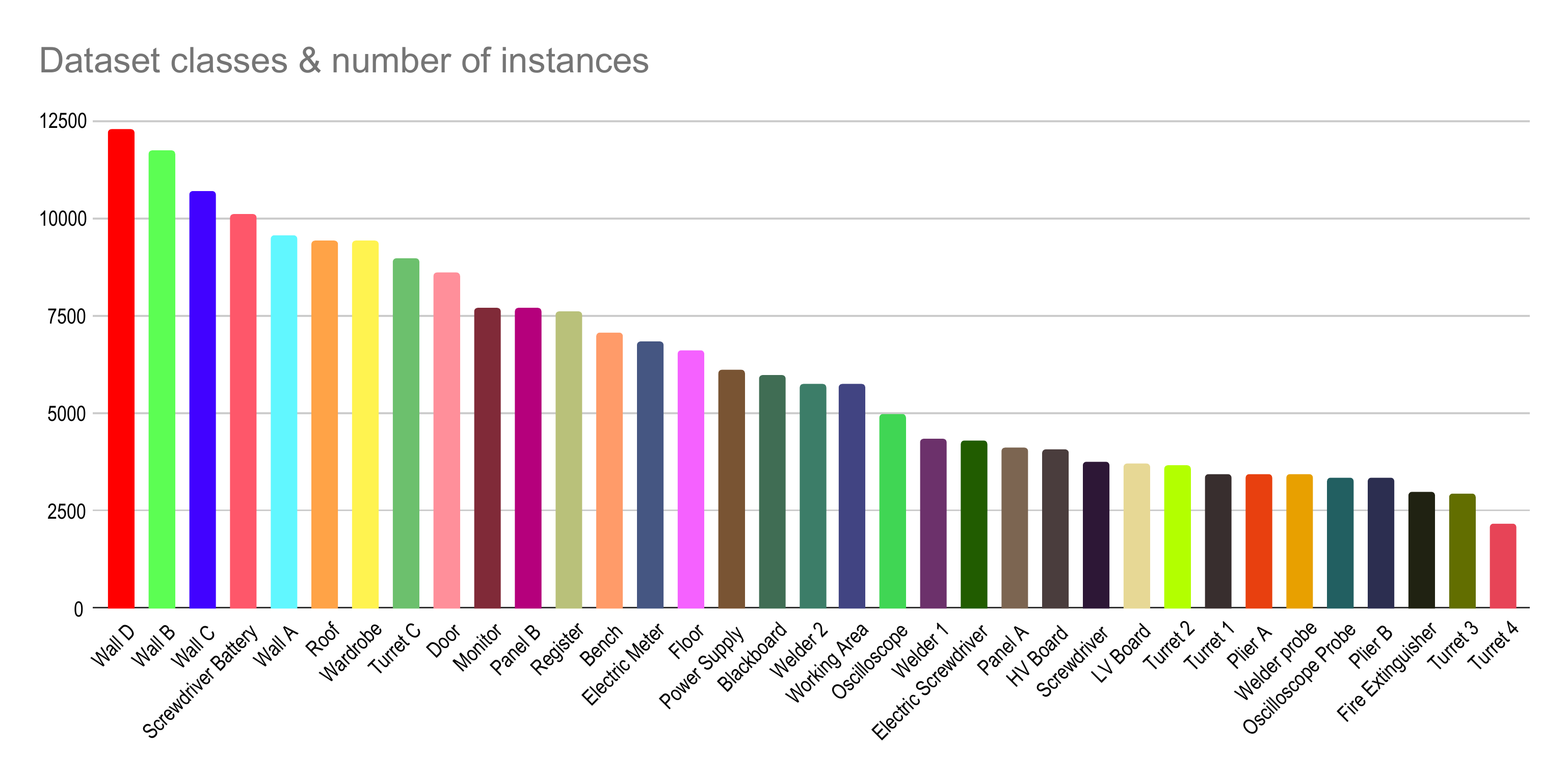}
            \caption{The classes of the dataset with the associated number of instances. The color of the bar also indicates the color associated to the class in the semantic image.} \label{classes}
        \end{figure}

    To study the considered problem, we have created a dataset comprised of two parts: real images with segmented masks manually annoted and synthetic images with automatically generated annotations. For each image, we have 4 annotations:
        \begin{description}
            \item [Semantic Label] Semantically annotated images, where each pixel is associated to a given semantic label (indicated by a specific color). We use a palette of 35 colors, each corresponding to a specific class (Fig.~\ref{classes} reports the classes along with the number of instances and associated colors of the synthetic dataset);
            \item [Panoptic Image] Annotated images which separate ``thing'' classes from ``stuff'' classes, assigning 0 to pixels belonging to the ``thing'' class and a contiguous id (from 0 to the n\textsuperscript{th} class) to pixels belonging to the ``stuff'' class. Unlabeled pixels are associated to the value 255. By ``thing'' we mean all those entities that can be counted (e.g. welder, piler, screwdriver, etc.), while the ``stuff'' class comprise those classes that represent amorphous regions of similar textures (e.g. walls, floor, etc.);
            \item[Instance Annotation] Textual annotation (a JSON in COCO format) that contains information about the segments within each image. This type of annotation also gives information about the bounding boxes and the class labels;
            \item[Panoptic Annotation] Textual annotation (a JSON in COCO format) that gives information on the area, class and bounding box of each instance. This annotation also contains an \textit{id} for each instance, calculated as $id = R + G \times 256 + B \times 256^2$, using the color of the belonging class as an RGB triplet.
        \end{description}
    After generating the semantic annotations (as described in the following paragraphs) instance annotations and panoptic annotations were generated from the semantic annotations computing the coordinates of the bounding boxes. Finally, panoptic images were generated using Detectron2~\footnote{\url{https://github.com/facebookresearch/detectron2}}.
    \begin{figure}[t]
        \centering
        \includegraphics[width=0.9\textwidth]{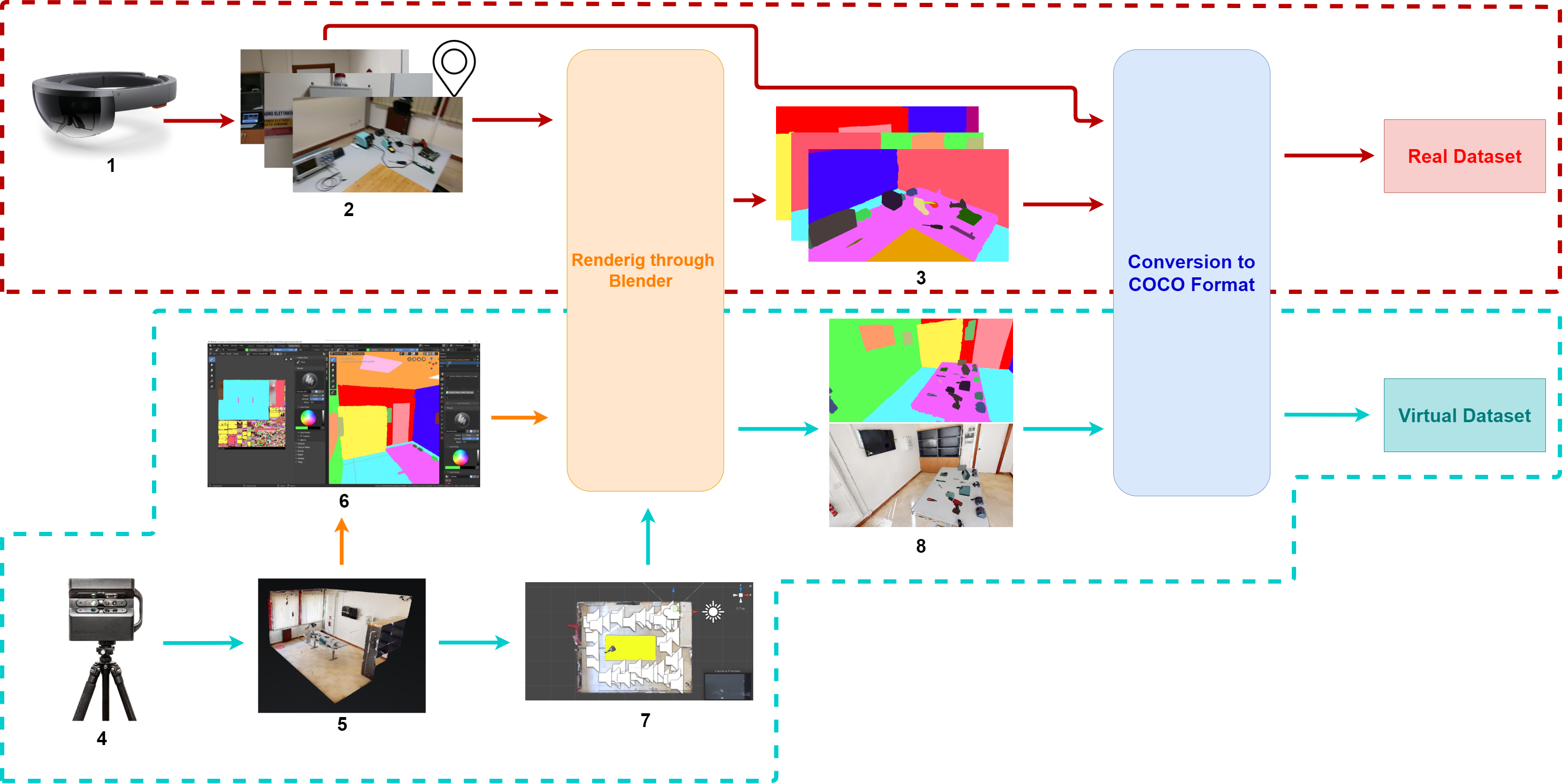}
        \caption{Dataset generation pipeline. \textbf{Red box} generation of the real dataset: (1) acquisition of real images using HoloLens2; (2) extraction of frames and related camera poses; (3) annotation of the segmentation masks. \textbf{Blue box} generation of the synthetic dataset: (4) acquisition of the 3D model using a Matterport 3D scanner; (5) generation of the 3D model; (6) semantic labelling of the 3D model using Blender; (7) generation of a random tour inside the 3D model; (8) generation of synthetic frames and semantic labels. \ (Rendering through Blender) the 3D model and the positions are processed by a script for the generation of frames and semantic labels; (Conversion in COCO format) semantic labels are processed by a script for extracting JSON annotations in COCO format.} \label{pipeline}
    \end{figure}
    
    \subsection{Real images acquisition and labeling}
\begin{figure}[t!]
            \centering
            \includegraphics[width=0.7\textwidth]{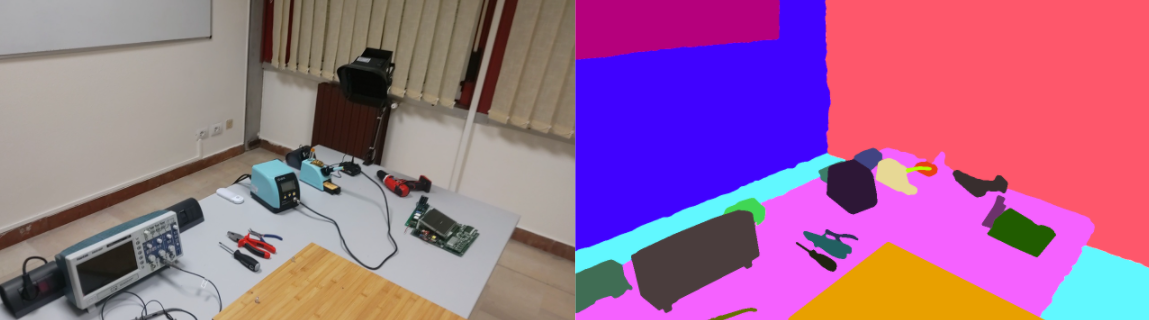}
            \caption{A real image containing  objects and corresponding segmentation mask.} \label{real_dataset}
        \end{figure}

    The real images contained in the dataset consist of 868 RGB images with a resolution of 1280x720 pixels. The images have been divided into a training set and a test set. Specifically, we used 668 images as a test and the remaining 200 images as a train. We choose an asymmetric split because we assume few real images are available for training, but we still want to have enough test images for evaluation.
    The acquisition of real images follows the pipeline shown in Fig.~\ref{pipeline} (red box). 
    The frames and corresponding camera poses, i.e. the position of the camera within the environment, were captured using a HoloLens2 device with a custom-made application which relies on HoloLens2 slam abilities to reconstruct accurate camera poses.
    All real images have been annotated for panoptic segmentation. To simplify the process, we use Blender in order to render semantic segmentation masks from the labeled 3D model of the scene (see next sections for details). The generated mask only contain static objects included in the 3D model. The movable objects present in the real images were annotated manually using the VGG Image Annotator (VIA)~\cite{dutta2019vgg} enabling to easily add the masks of the mobile objects to the automatically generated segmentation masks (Fig.~\ref{real_dataset}). 
    
    \subsection{Synthetic images generation}
    
         \begin{figure}[t!]
         \centering
          \includegraphics[width=0.7\textwidth]{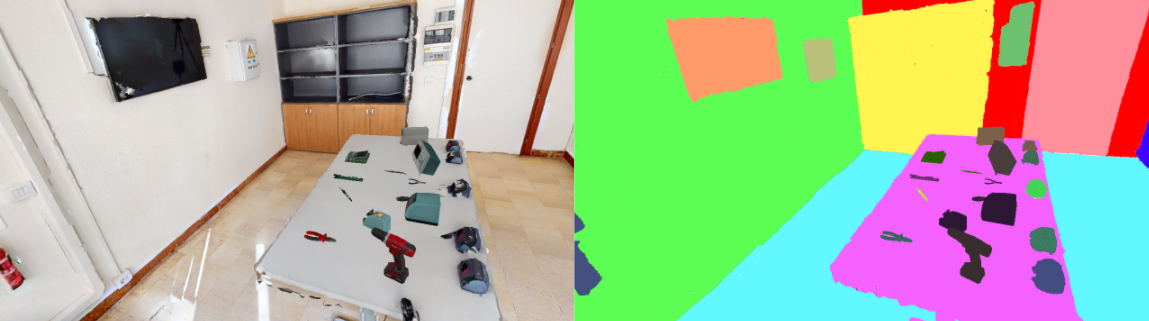}
            \caption{A synthetic image and the corresponding segmentation mask.} \label{synthetic_dataset}
        \end{figure}
        
        The generation of the synthetic images follows the pipeline shown in Fig.~\ref{pipeline} (blue box). This part of the dataset consists of 25,079 RGB images with a resolution of 1280x720 pixels. The images have been divided into a training set, a test set and a validation set following a 70:20:10 proportion (see an example in Fig.~\ref{synthetic_dataset}). 
    
       The first step was to acquire the 3D model of the environment using the Matterport 3D scanner~\footnote{\url{https://matterport.com/}}. Once the 3D model was obtained, a tool~\cite{orlando2019image} developed with Unity3D~\footnote{\url{http://unity3d.com/unity/}} was used to simulate an agent that navigates within the 3D space varying the position and rotation of the camera in order to acquire images from several positions and points of view. The position and rotation of the camera are automatically obtained from the position of the the agent's head.
        
        The 3D model obtained through the Matterport scan was fed to Blender~\footnote{\url{https://www.blender.org/}} in order to obtain the semantic labels by manual labelling of the 3D model. In this stage, using the Blender brush tool, the 3D model was colored accordingly to the considered classes showed in Fig.~\ref{classes} to obtain a fine-grained semantic label of the overall model. This labeling has been be used to automatically generate semantic images from given positions within the model aligned with the synthetic rgb images.

        \subsubsection{Movable objects scanning and labeling}
\begin{figure}[t!]
            \centering
            \includegraphics[width=0.65\textwidth]{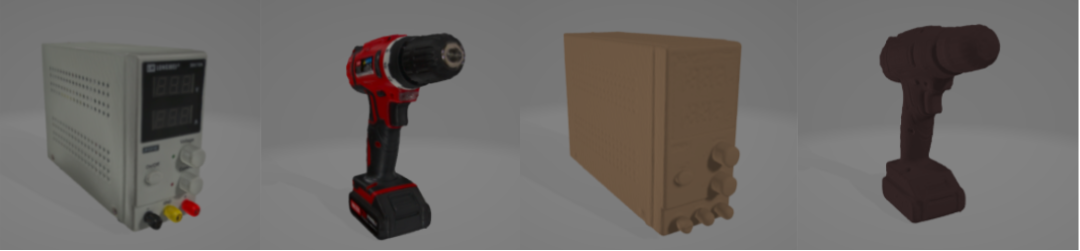}
            \caption{Two examples of the acquired movable items in the first two images and the corresponding monochrome textures in the last two images.} \label{object}
        \end{figure}

        To model the presence of movable objects positioned on the workbench (e.g. power supply, oscilloscope, clamp, etc.), a 3D model of each object was first acquired using an Artec scanner~\footnote{\url{https://www.artec3d.com/}}. Photorealism was obtained during post processing: the real texture was preserved and a semantic monochrome texture was generated for the semantic label (Fig.~\ref{object}).
        
        The scanned objects were inserted into the 3D model on Blender, with the constraint of being over the workbench. Finally, annotated frames were extracted through two indipendent renderings in Blender: one with the original textures and one with the semantic textures (Fig.~\ref{synthetic_dataset}). During data generation, the position of the objects on the workbench has been randomly changed every 5 frames.

    \begin{table}[t!]
\centering
\resizebox{0.8\columnwidth}{!}{
\begin{tabular}{r|ccccccccc}
             
              \multicolumn{10}{c}{\textbf{Panoptic Quality [PQ]}}                                                                                             \\ \hline
             \textbf{\# of Real Training Img.}    & \textbf{0}           & \textbf{25} & \textbf{50} & \textbf{75} & \textbf{100} & \textbf{125} & \textbf{150} & \textbf{175} & \textbf{200} \\\hline
\textbf{Training with R Img.}   & \multicolumn{1}{l}{} & 8.95        & 13.48       & 14.55       & 15.04        & 15.58        & 16.04        & 16.75        & 17.72        \\
\textbf{Training with R+S Img.} & \textbf{15.88}                & \textbf{16.30}        & \textbf{19.55}       & \textbf{20.05}       & \textbf{20.49}        & \textbf{20.63}        & \textbf{20.8}         & \textbf{21.19}        & \textbf{22.38}        \\ \hline 
              \multicolumn{10}{c}{\textbf{Segmentation Quality [SQ]}}                                                                                         \\ \hline
            \textbf{\# of Real Training Img.}  & \textbf{0}           & \textbf{25} & \textbf{50} & \textbf{75} & \textbf{100} & \textbf{125} & \textbf{150} & \textbf{175} & \textbf{200} \\\hline
\textbf{Training with R Img.}   & \multicolumn{1}{l}{} & 25.57       & 40.02       & 44.05       & 40.44        & 41.33        & 41.97        & 42.58        & 42.93        \\
\textbf{Training with R+S Img.}  & \textbf{49.15}                & \textbf{43.36}       & \textbf{44.95}       & \textbf{46.03}       & \textbf{47.10}         & \textbf{48.75}        & \textbf{48.91}        & \textbf{49.17}        & \textbf{49.47}        \\ \hline 
              \multicolumn{10}{c}{\textbf{Recognition Quality [RQ]}}                                                                                          \\ \hline
            \textbf{\# of Real Training Img.}  & \textbf{0}           & \textbf{25} & \textbf{50} & \textbf{75} & \textbf{100} & \textbf{125} & \textbf{150} & \textbf{175} & \textbf{200} \\\hline
\textbf{Training with R Img.}     & \multicolumn{1}{l}{} & 12.03       & 18.39       & 20.15       & 20.45        & 21.65        & 22.13        & 23.01        & 24.36        \\
\textbf{Training with R+S Img.}  & \textbf{22.58}                & \textbf{23.36}       & \textbf{27.46}       & \textbf{28.14}       & \textbf{28.6}         & \textbf{28.67}        & \textbf{29.16}        & \textbf{29.55}        & \textbf{31.69}        \\ \hline 
             \multicolumn{10}{c}{\textbf{Average Precision Bounding Box [AP (bbox)]}}                                                                                                    \\ \hline 
             \textbf{\# of Real Training Img.} & \textbf{0}           & \textbf{25} & \textbf{50} & \textbf{75} & \textbf{100} & \textbf{125} & \textbf{150} & \textbf{175} & \textbf{200} \\\hline
\textbf{Training with R Img.}     & \multicolumn{1}{l}{} & 11.99       & 23.10        & 26.33       & 27.46        & 27.68        & 27.79        & 27.81        & 28.82        \\
\textbf{Training with R+S Img.}  & \textbf{22.16}                & \textbf{28.86}       & \textbf{34.42}       & \textbf{38.59}       & \textbf{39.86}        & \textbf{39.91}        & \textbf{41.81}        & \textbf{42.57}        & \textbf{43.56}        \\ \hline 
              \multicolumn{10}{c}{\textbf{Average Precision Segmentation [AP (seg)]}}                                                                                                    \\ \hline 
             \textbf{\# of Real Training Img.} & \textbf{0}           & \textbf{25} & \textbf{50} & \textbf{75} & \textbf{100} & \textbf{125} & \textbf{150} & \textbf{175} & \textbf{200} \\\hline
\textbf{Training with R Img.}     & \multicolumn{1}{l}{} & 10.50        & 20.54       & 23.35       & 24.68        & 25.30         & 26.72        & 27.93        & 26.25        \\
\textbf{Training with R+S Img.}  & \textbf{19.84}                & \textbf{26.58}       & \textbf{32.49}       & \textbf{34.47}       & \textbf{34.51}        & \textbf{36.14}        & \textbf{37.84}        & \textbf{38.13}        & \textbf{39.05}        \\ \hline 
              \multicolumn{10}{c}{\textbf{Mean Intersection over Union [mIoU]}}                                                                                                         \\ \hline
            \textbf{\# of Real Training Img.}  & \textbf{0}           & \textbf{25} & \textbf{50} & \textbf{75} & \textbf{100} & \textbf{125} & \textbf{150} & \textbf{175} & \textbf{200} \\\hline
\textbf{Training with R Img.}     & \multicolumn{1}{l}{} & 40.25       & 60.88       & 69.47       & 74.47        & 75.11        & 76.38        & 77.10         & 77.93        \\
\textbf{Training with R+S Img.}  & \textbf{55.43}                & \textbf{66.63}       & \textbf{71.67}       & \textbf{78.90}        & \textbf{80.84}        & \textbf{82.95}        & \textbf{83.06}        & \textbf{83.29}        & \textbf{84.39} \\
\hline
\end{tabular}%
}
\caption{Results of Panoptic Segmentation with real (R) and synthetic (S) data. }
\label{table:results_table}
\end{table}

\section{Experimental Settings and Results}
\label{benchmark}
    We perform experiments to evaluate a panoptic segmentation model trained on synthetic data and finetuned on real images considering a variable amount of real data, i.e. 200 images in our case. The experiments compare two different approaches that use synthetic and real data.
    \begin{description}
        \item [Real data (R)] In this experiment, we train the model using only real data. To assess the amount of real data needed to obtain good performances, the model was trained with different amounts of real data, and specifically 25, 50, 75, 100, 125, 150, 175, and 200 images.
        \item [Real and Synthetic data (R+S)] In this experiment, we uses both real and synthetic data to train the model. The training procedure consists of two steps: in the first one, we train the model using only synthetic data; in the second one, we fine-tune the model with real data. As in the aforementioned experiment, the amount of real data varies between 25 and 200 images. We also include the case in which no real data is used for fine-tuning, and hence the model is only trained on synthetic data.
    \end{description}
        The metrics used for model evaluation are the classic metrics used for the instance segmentation and semantic segmentation tasks, namely Intersection-Over-Union (IoU)~\cite{rezatofighi2019generalized} and Average Precision (AP)~\cite{10.1007/978-3-319-10602-1_48}.
        The results of the experiments in terms of these metrics are used to calculate the Panoptic Quality (PQ)~\cite{Kirillov_2019_CVPR} using two terms: Segmentation Quality (SQ) and Recognition Quality (RQ). 
        
    \begin{center}
            \begin{figure}[t!]
        \begin{subfigure}{.5\textwidth}
          \centering
          
          \includegraphics[width=0.9\linewidth]{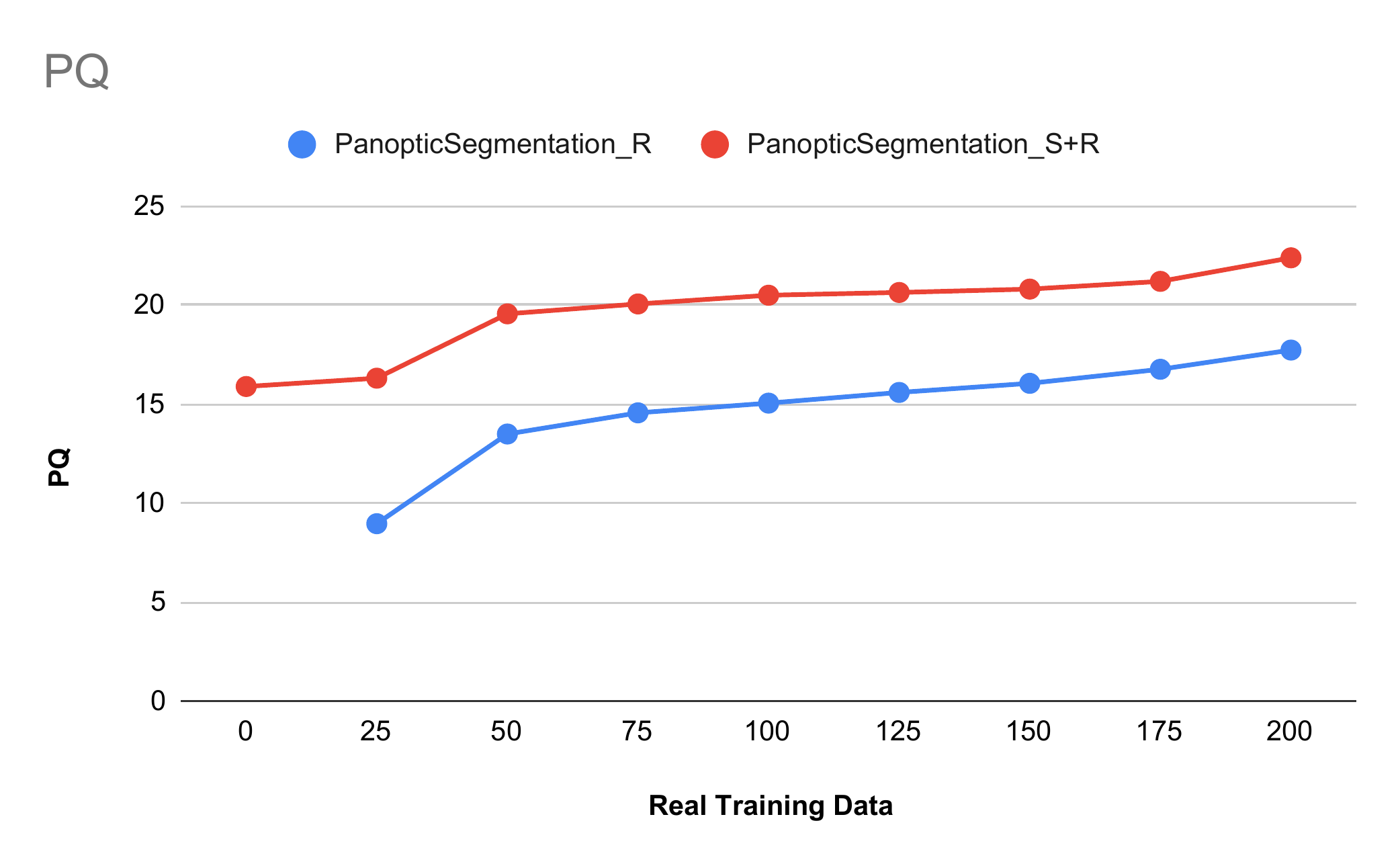}  
          \caption{}
          \label{fig:sub-first}
        \end{subfigure}
        \begin{subfigure}{.5\textwidth}
          \centering
          
          \includegraphics[width=0.9\linewidth]{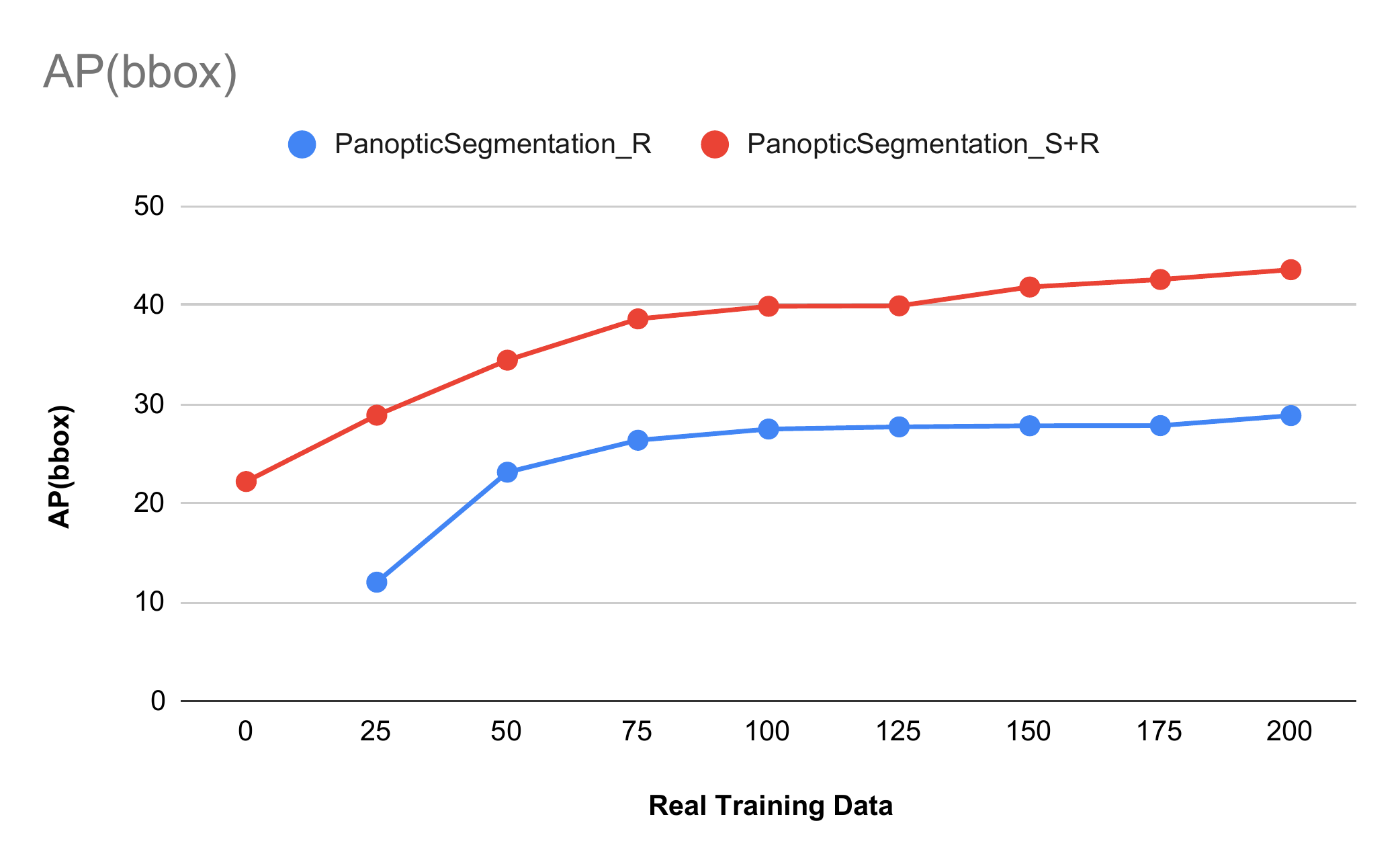}  
          \caption{}
          \label{fig:sub-second}
        \end{subfigure}
        \newline
        \begin{subfigure}{.5\textwidth}
          \centering
          
          \includegraphics[width=0.9\linewidth]{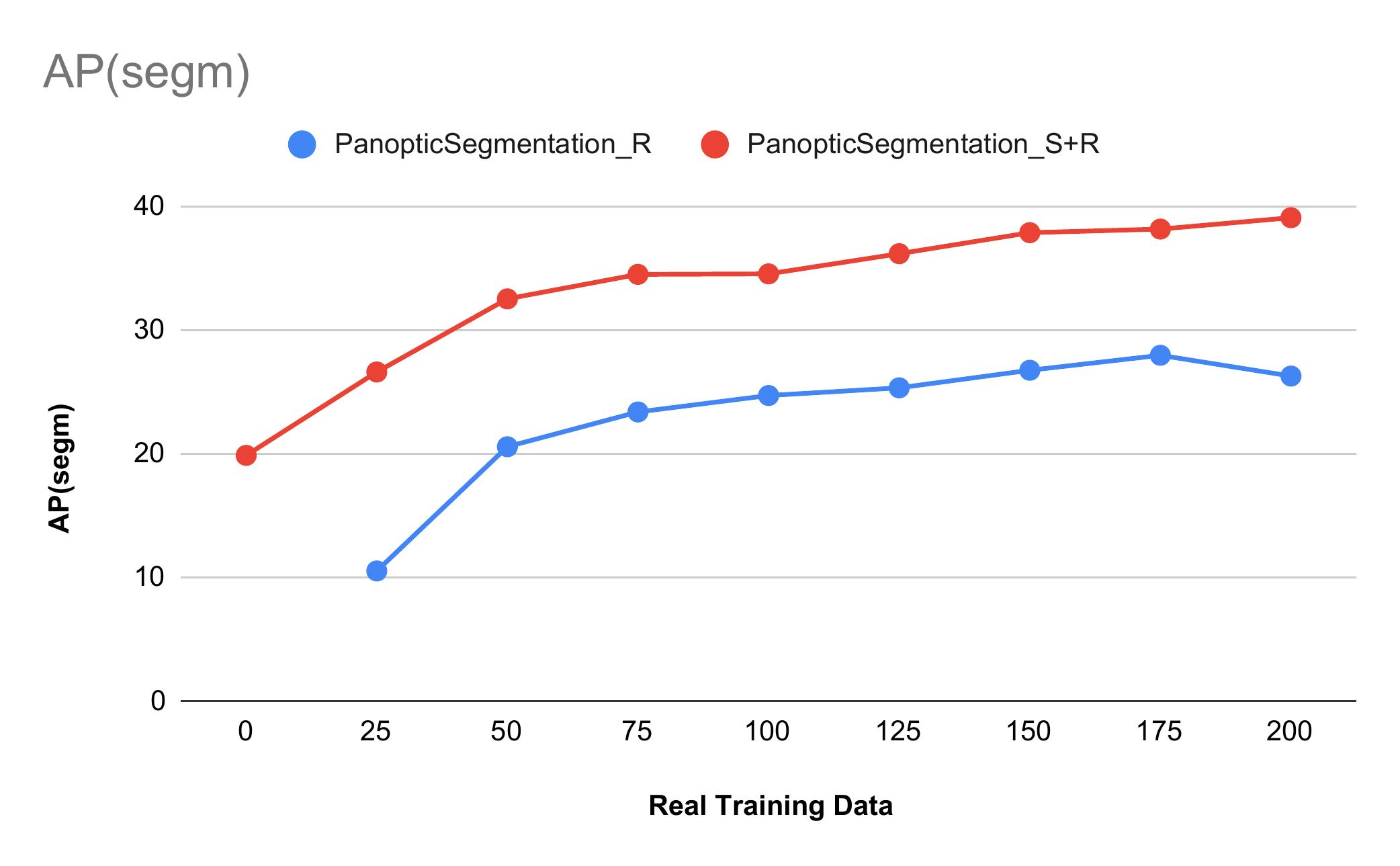}  
          \caption{}
          \label{fig:sub-third}
        \end{subfigure}
        \begin{subfigure}{.5\textwidth}
          \centering
          
          \includegraphics[width=0.9\linewidth]{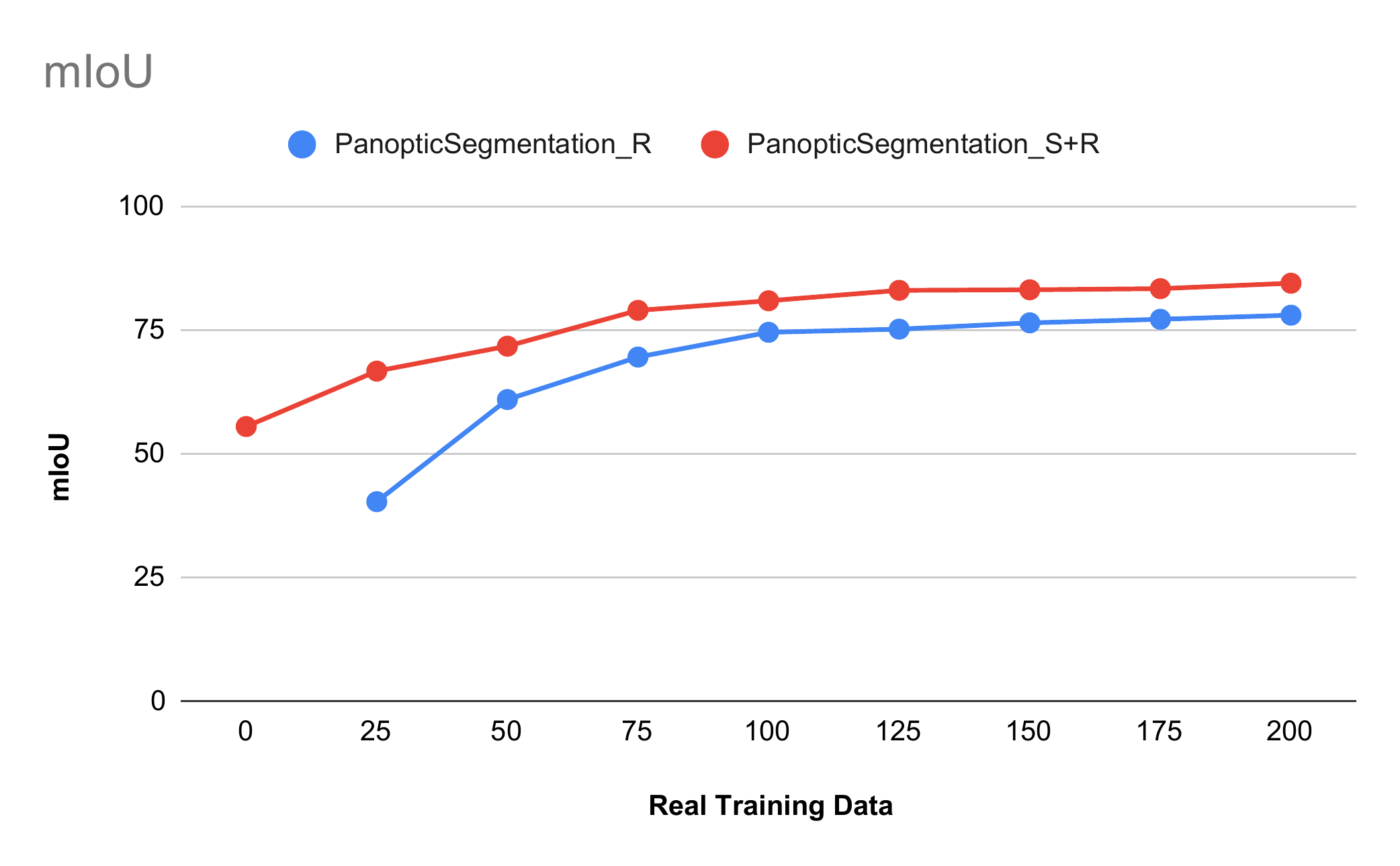}  
          \caption{}
          \label{fig:sub-fourth}
        \end{subfigure}
        \caption{Trends of four metrics for different quantities of real data.}
        \label{linegraph}
        \end{figure}
        \end{center}
        In Table \ref{table:results_table}, we report the results of the experiments.
        The numbers show that the use of synthetic data is key to reduce the need of real images. For instance, in the case of the Panoptic Quality, using only synthetic images performs better than using only 125 real images (15.88\% vs 15.58\%), and using synthetic images plus 50 real images is better than using only 200 real images (19.55\% vs 17.72\%). This allows to obtain significant savings in the amount of human work needed to label images, which translates also to a faster deploy. Similar observations apply to all the measures except for mIoU, in which case the real images needed to overcome results obtained with 200 real images are only 75 (78.9\% vs 77.93\%).
        
\begin{figure}[t!]
        \centering
          \includegraphics[width=0.7\textwidth]{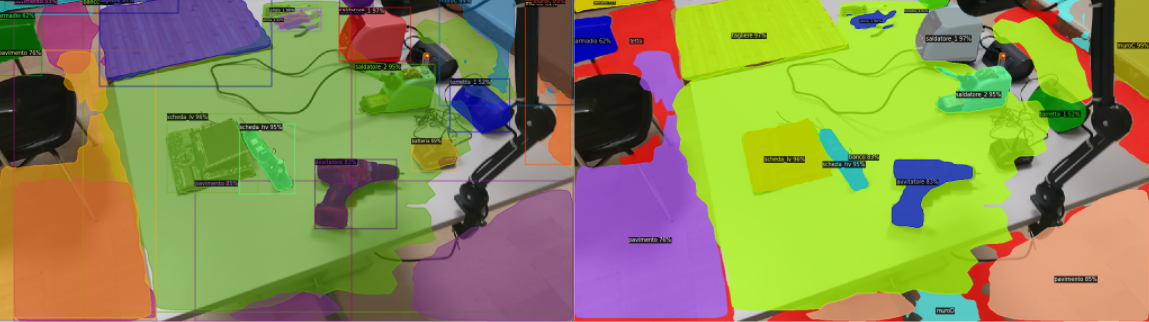}
            \caption{An example of qualitative result obtained with the best performing model (Synthetic + 200 real images). On the right, the result of the Instance Segmentation task, whereas, on the left, the result of Semantic Segmentation.} \label{qualitative}
        \end{figure}

        Fig.~\ref{linegraph}(a)-(d) summarize the results graphically for four metrics. The curves show that PanopticSegmentation\_S+R achieves much better results using the same amounts of real data than the baseline PanopticSegmentation\_R. Fig.~\ref{qualitative} shows a qualitative result generated using the best performing model.

\section{Conclusion}
\label{conclusion}
    In this work, we have considered the problem of training a panoptic segmentation model in an industrial environment exploiting a large amount of synthetic data and a small set of real images. We presented a pipeline to generate semantically labeled synthetic data and proposed a dataset suitable to study the problem. We hence benchmarked a panoptic segmentation baseline on the proposed data in the presence of all synthetic images and varying amounts of real images. The results show that the use of synthetic data is key to bootstrap the training of the model, decreasing the cost of manual labeling, which speeds up the deploy.

\section*{Acknowledgements}
This research is supported by Next Vision\footnote{Next Vision: \href{https://www.nextvisionlab.it/}{https://www.nextvisionlab.it/}} s.r.l., and the project MEGABIT - PIAno di inCEntivi per la RIcerca di Ateneo 2020/2022 (PIACERI) – linea di intervento 2, DMI - University of Catania.

\clearpage

\bibliographystyle{splncs04}
\bibliography{refs}

\end{document}